\documentclass[sigconf]{acmart}
\pdfoutput=1
\usepackage{booktabs} % For formal tables
\usepackage{overpic}
\usepackage{amsmath}
\usepackage{graphicx}
\usepackage{dblfloatfix}
\usepackage{subcaption}
\usepackage{balance}
\setcopyright{rightsretained}
\author{Yuanzheng Ci}
\affiliation{%
	\department{DUT-RU International School of Information Science \& Engineering}
	\institution{Dalian University of Technology}
}
\email{orashi@mail.dlut.edu.cn}

\author{Xinzhu Ma}
\affiliation{%
	\department{DUT-RU International School of Information Science \& Engineering}
	\institution{Dalian University of Technology}
}
\email{maxinzhu@mail.dlut.edu.cn}

\author{Zhihui Wang}\authornote{Corresponding author.}\authornote{Also with DUT-RU International School of Information Science \& Engineering, Dalian University of Technology.\vspace{-1pt}}
\affiliation{%
	\department{Key Laboratory for Ubiquitous Network and Service Software of Liaoning Province}
	\institution{Dalian University of Technology}
}
\email{zhwang@dlut.edu.cn}
\author{Haojie Li$^\dag$}
\affiliation{
	\department{Key Laboratory for Ubiquitous Network and Service Software of Liaoning Province}
	\institution{Dalian University of Technology}
}
\email{hjli@dlut.edu.cn}
\author{Zhongxuan Luo$^\dag$}
\affiliation{
	\department{Key Laboratory for Ubiquitous Network and Service Software of Liaoning Province}
	\institution{Dalian University of Technology}
}
\email{zxluo@dlut.edu.cn}

\begin{document}
	
\title{User-Guided Deep Anime Line Art Colorization with Conditional Adversarial Networks}

\begin{abstract}
Scribble colors based line art colorization is a challenging computer vision problem since neither greyscale values nor semantic information is presented in line arts, and the lack of authentic illustration-line art training pairs also increases difficulty of model generalization.  Recently, several Generative Adversarial Nets (GANs) based methods have achieved great success. They can generate colorized illustrations conditioned on given line art and color hints. However, these methods fail to capture the authentic illustration distributions and are hence perceptually unsatisfying in the sense that they often lack accurate shading. 
To address these challenges, we propose a novel deep conditional adversarial architecture for scribble based anime line art colorization.  Specifically, we integrate the conditional framework with WGAN-GP criteria as well as the perceptual loss to enable us to robustly train a deep network that makes the synthesized images more natural and real. We also introduce a local features network that is independent of synthetic data. With GANs conditioned on features from such network, we notably  increase the generalization capability over "in the wild" line arts. Furthermore, we collect two datasets that provide high-quality colorful illustrations and authentic line arts for training and benchmarking. With the proposed model trained on our illustration dataset, we demonstrate that images synthesized by the presented approach are considerably more realistic and precise than alternative approaches.

\end{abstract}

%
% The code below should be generated by the tool at
% http://dl.acm.org/ccs.cfm
% Please copy and paste the code instead of the example below.
%

\copyrightyear{2018} 
\acmYear{2018} 
\setcopyright{acmcopyright}
\acmConference[MM '18]{2018 ACM Multimedia Conference}{October 22--26, 2018}{Seoul, Republic of Korea}
%\acmBooktitle{2018 ACM Multimedia Conference (MM '18), October 22--26, 2018, Seoul, Republic of Korea}
\acmPrice{15.00}
\acmDOI{10.1145/3240508.3240661}
\acmISBN{978-1-4503-5665-7/18/10}




\maketitle

\begin{figure}[t]
	\centering
	\includegraphics[width=0.48\textwidth]{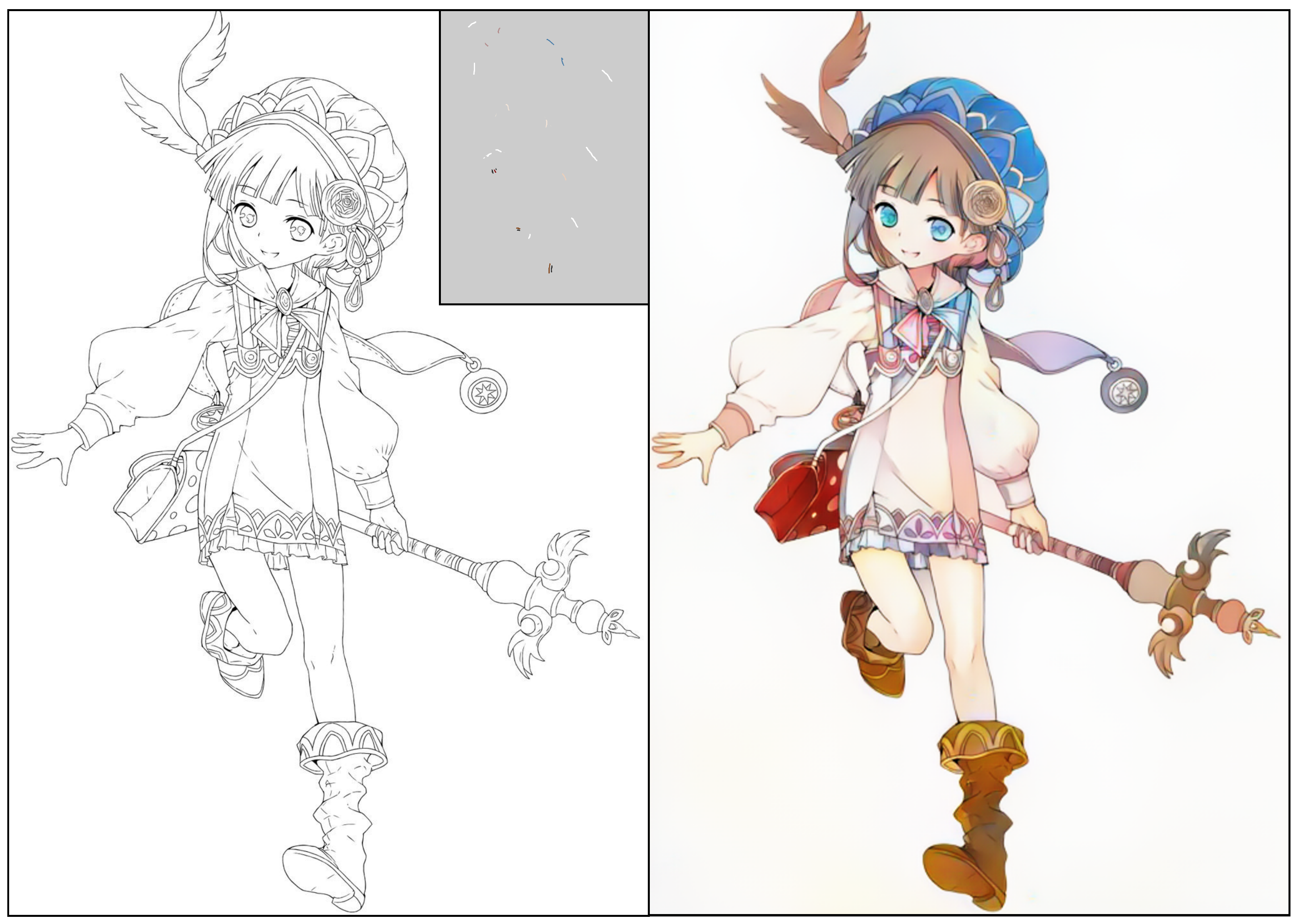}
	\caption{Our proposed method colorizes a line art composed by artist (left) based on guided stroke colors (top center, best viewed with grey background) and learned priors. Line art image is from our collected line art dataset. }
	\label{fig:title}
\end{figure}

\section{Introduction}

Line art colorization plays a critical role in the workflow of artistic work such as the composition of illustration and animation. Colorizing the in-between frames of animation as well as  simple illustrations involves a large portion of redundant works. Nevertheless, there is no automatic colorization pipelines for line art colorization. As a consequence, it is performed manually using image editing applications such as Photoshop and PaintMan. Especially in the animation industry, where it is even known as a hard labor. It's a challenging task to develop a fast and straightforward way to produce illustration-realistic imagery with line arts.

Several recent methods have explored approaches for guided image colorization \cite{sangkloy2017scribbler, zhang2017real, hensman2017cgan, furusawa2017comicolorization, paintschainer, zhang2017style, liu2017auto, frans2017outline}. Some works are mainly focused on images containing greyscale information \cite{zhang2017real, furusawa2017comicolorization, hensman2017cgan}. A user can colorize a greyscale image by color points or color histograms \cite{zhang2017real}, or colorize a manga based on reference images \cite{hensman2017cgan} with color points \cite{furusawa2017comicolorization}. These methods can achieve impressive results but can neither handle sparse input like line arts, nor color stroke-based input which is easier for users to intervene. To address these issues, recently,  researchers have also explored more data-driven colorization methods \cite{sangkloy2017scribbler, liu2017auto, frans2017outline}. These method colorize a line art/sketch by scribble color strokes based on learned priors over synthetic sketches. This makes colorizing a new sketch cheaper and easier. Nevertheless, the results often contains unrealistic colorization and artifacts. More fundamentally, the overfitting over synthetic data is not well handled, thus the network can hardly perform well with "in the wild" data. PaintsChainer\cite{paintschainer} tries to address these issues with three models (with two models not opened) proposed. However, the eye pleasing results were achieved in the cost of losing realistic textures  and  global shading that authentic illustrations preserve.

In this paper, we propose a novel conditional adversarial illustration synthesis architecture trained fully on synthetic line arts. Unlike existing approaches using typical conditional networks \cite{isola2017image}, we combine a cGAN with a pretrained local features network, i.e., both generator and discriminator network are only conditioned on the output of local features network to increase the generalization ability over authentic line arts. With proposed networks trained in an end-to-end manner with synthetic line arts, we are able to  generate illustration-realistic colorization from sparse, authentic line art boundaries and color strokes. This means we do not suffer from overfitting issue over synthetic data. 
In addition, we randomly simulate user interactions during training,  allowing the network to propagate the sparse stroke colors to semantic scene elements. Inspired by \cite{kupyn2017deblurgan} which obtained state-of-the art results in motion deblurring, we fuse the conditional framework \cite{isola2017image} with WGAN-GP \cite{gulrajani2017improved} and perceptual loss \cite{johnson2016perceptual} as the criterion  in the GAN training stage. This allows us to robustly train a network with more capacity, thus makes the synthesized images more natural and real. Moreover, we collected two cleaned datasets with high quality color illustrations  and hand-drawn line arts. They provide a stable training data source as well as a test benchmark for line art colorization.

By training with the proposed illustration dataset and adding minimal augmentation, our model can handle general anime line arts with stroke color hints. As the loss term optimizes the results to resemble the ground truth, it mimics realistic color allocations and general shadings with respect to the color strokes and scene elements as shown in Figure \ref{fig:title}. We trained our model with millions of image pairs, and achieve significant improvements in stroke-guided line art colorization. Extensive experimental results demonstrate that the performance of the proposed method is far superior to those of the state-of-the-art stroke-based user-guided line art colorization methods in both qualitative and quantitative evaluations.

In summary, the key contributions of this paper are summarized as follows.

\begin{itemize}
	\item We propose an illustration synthesis architecture and a loss for stroke-based user-guided anime line art colorization, whose results are significantly better than existing guided line art colorization methods.
	\item We introduce a novel local features network in the cGAN architecture to enhance the generalization ability of the networks trained with synthetic data.
	\item  The colorization network in our cGAN is different with existing GANs' generators in that it is much deeper, with several specially designed layers to  both increase the receptive  field and the network capacity. It makes the synthesized images more natural and real.
	\item We collect two datasets that provide quality illustration training data and  line art test benchmark.
\end{itemize}
\section{Related Work}
\subsection{User-guided colorization}
Early interactive colorization methods \cite{levin2004colorization, huang2005adaptive}  propagate stroke colors with low-level similarity metrics. These methods based on the assumption that the adjacent pixels with similar luminousness in greyscale images should have a similar color, and numerous user interactions are typically required to achieve realistic colorization results. Later research studies improved and extended this method by using chrominance blending \cite{yatziv2006fast}, specific schemes for different textures \cite{qu2006manga}, better similarity metrics \cite{luan2007natural}  and global optimization with all-pair constraints \cite{an2008appprop, xu2009efficient}. Learning  methods such as boosting \cite{li2008scribbleboost}, manifold learning \cite{chen2012manifold} and neural networks \cite{endo2016deepprop, zhang2017real} have also been proposed to propagate stroke colors with learned priors. In addition to local control, some approaches proposed to colorize images by transferring the color theme \cite{li2015image, furusawa2017comicolorization, hensman2017cgan} or color palette \cite{chang2015palette, zhang2017real} of the reference image. While these methods make use of the greyscale information of the source image which is not available for line art/sketch,  Scribbler \cite{Sangkloy2016Scribbler} developed a system to transform  sketches of specific categories to real images with scribble color strokes. Frans \cite{frans2017outline} and Liu et al. \cite{liu2017auto} proposed methods for guided line art colorization but can hardly produce plausible results based on arbitrary man-made line arts. Concurrently, Zhang et al. \cite{zhang2017style} colorize man-made anime line arts with a reference image. PaintsChainer \cite{paintschainer} first developed an online application that can generate pleasing colorization results for man-made anime line arts  with stroke colors as hints, they provide three models (named \textit{tanpopo}, \textit{satsuki}, \textit{canna}) with one of them open-sourced (\textit{tanpopo}). However, these models failed to capture the authentic illustration distribution and thus lack of accurate shading. 
\subsection{Automatic colorization}
Recently, colorization methods that do not require color information were proposed \cite{cheng2015deep, deshpande2015learning, iizuka2016let, zhang2016colorful}. These methods train CNNs \cite{lecun1998gradient} on large datasets to learn a direct mapping from greyscale images to colors. The learning based methods can combine the low-level details as well as high-level semantic information to produce photo-realistic colorization that perceptually pleasing the people. In addition, Isola et al. \cite{isola2017image}, Zhu et al. \cite{zhu2017unpaired} and Chen et al. \cite{chen2018sketchygan}  learn a direct mapping from human drawn sketches (for a particular category or with category labels) to  realistic images with generative adversarial networks. Larsson et al. \cite{larsson2016learning} and Guadarrama et al. \cite{guadarrama2017pixcolor} also provides solutions to the multi-modal uncertainty of the colorization problem as their methods can also generate multiple results. However limitations still exist as they can only cover a small subset of the possibilities. Beyond learning a direct mapping,  Sketch2Photo \cite{chen2009sketch2photo} and PhotoSketcher \cite{eitz2011photosketcher} synthesize realistic images by compositing objects and backgrounds retrieved from a large collection of images based on a given sketch.
\subsection{Generative Adversarial Networks}
Recent study of GANs \cite{goodfellow2014generative, radford2015unsupervised} has achieved great success in a wide range of image synthesis applications, including blind motion deblurring \cite{nah2017deep, kupyn2017deblurgan}, high-resolution image syhthesis \cite{wang2017high, karras2017progressive}, photo-realistic super-resolution \cite{ledig2016photo}  and image in-paining \cite{pathak2016context}. The GAN training strategy is to define a game between two competing networks. The generator attempts to fool a simultaneously trained discriminator that classifies images as real or synthetic. GANs are known for its ability to generate samples of good perceptual quality, however, the vanilla version of GAN suffers from many problems such as mode collapse, vanishing gradients etc, as described in \cite{salimans2016improved}. Arjovsky et al. \cite{arjovsky2017wasserstein} discuss the difficulties in GAN training caused by the vanilla loss function and propose to use the approximation of Earth-Mover (also called Wasserstein-1) distance as the critic. Gulrajani et al. \cite{gulrajani2017improved} further improved its stability with gradient penalty thus enable us to train more architectures with almost no hyperparameter tuning. The basic GAN framework can also be augmented using side information. One strategy is to supply both the generator and discriminator with class labels to produce class conditional samples, which is known as cGAN \cite{mirza2014conditional}. This kind of side information can significantly improve the quality of generated samples \cite{van2016conditional}. Richer side information such as paired input images \cite{isola2017image},  boundary map \cite{wang2017high} and image captions \cite{reed2016generative} can improve sample quality further. However, when training data has different patterns compared with test data (in our case, the synthetic line art and authentic line art), existing frameworks cannot perform reasonably well.

\section{Proposed Method}
\begin{figure}[t]
	\centering
	\includegraphics[width=0.48\textwidth]{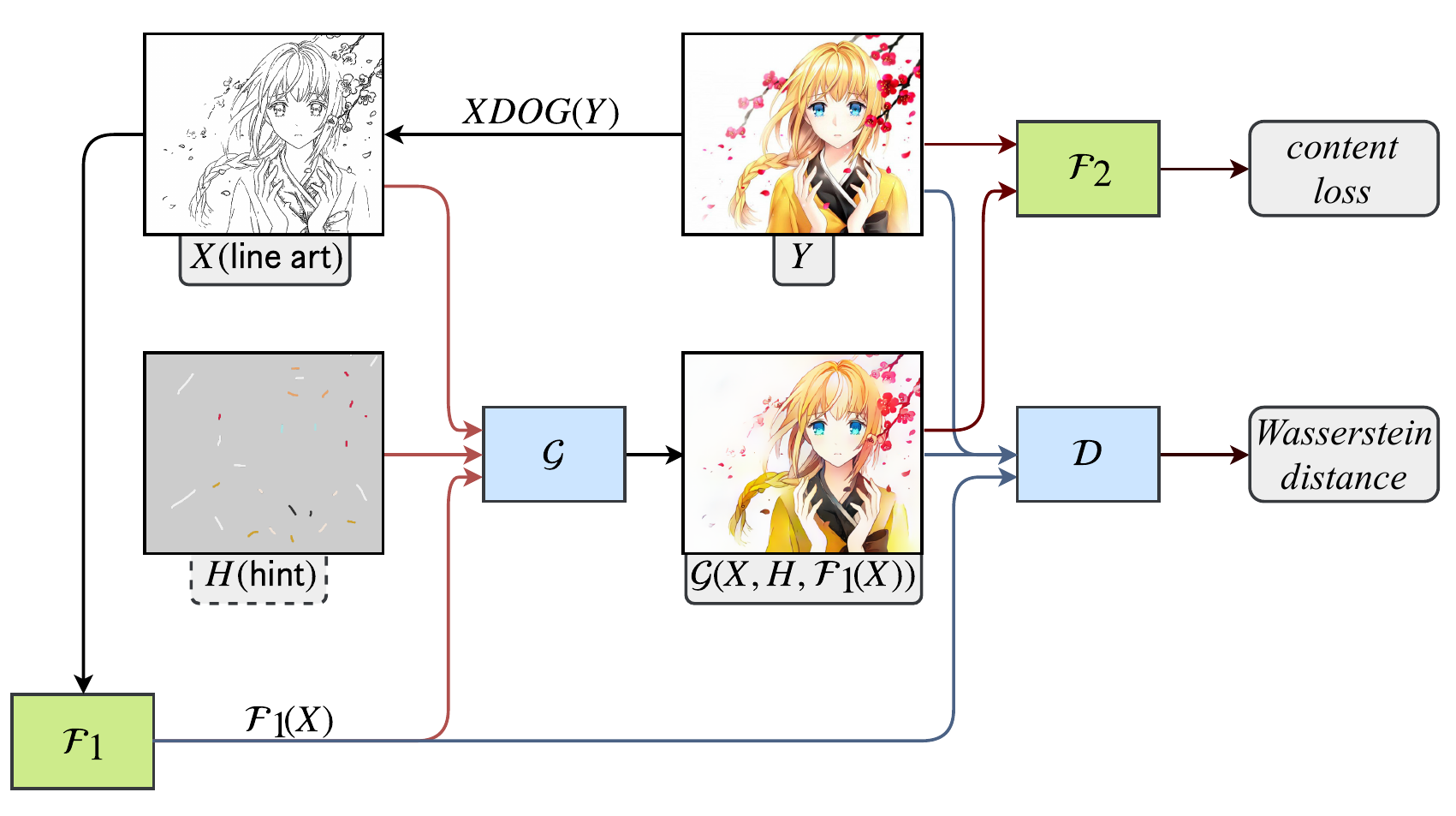}
	\caption{Overview of our cGAN based colorization model. The training proceeds with feature extractor  ($\mathcal{F}_{1}$, $\mathcal{F}_{2}$), generator ($\mathcal{G}$)  and discriminator ($\mathcal{D}$) to help $\mathcal{G}$ learns to generate colorized image $\mathcal{G}(X,H,\mathcal{F}_{1}(X))$ based on line art image $X$ and color hint $H$. Network $\mathcal{F}_{1}$ extracts semantical feature maps from $X$, while we do not feed $X$ to $\mathcal{D}$ to avoid being overfitted on the characteristic of synthetic line arts. Network $\mathcal{D}$ learns to give a wasserstein distance between $\mathcal{G}(X,H, \mathcal{F}_{1}(X))$-$\mathcal{F}_{1}(X)$ pairs and $Y$-$\mathcal{F}_{1}(X)$ pairs.}
	\label{fig:overview}
	
\end{figure}
Given line arts and user inputs, we train a deep network to synthesis illustrations. In Section \ref{sec:learn}, we introduce the objective of our network. We then describe the loss functions of our system in Section \ref{sec:loss}. In Section \ref{sec:network}, we define our network architecture. Finally we describe the user interaction mechanism in Section \ref{sec:interact}.
\begin{figure*}[ht]
	\centering
	\includegraphics[width=1\textwidth]{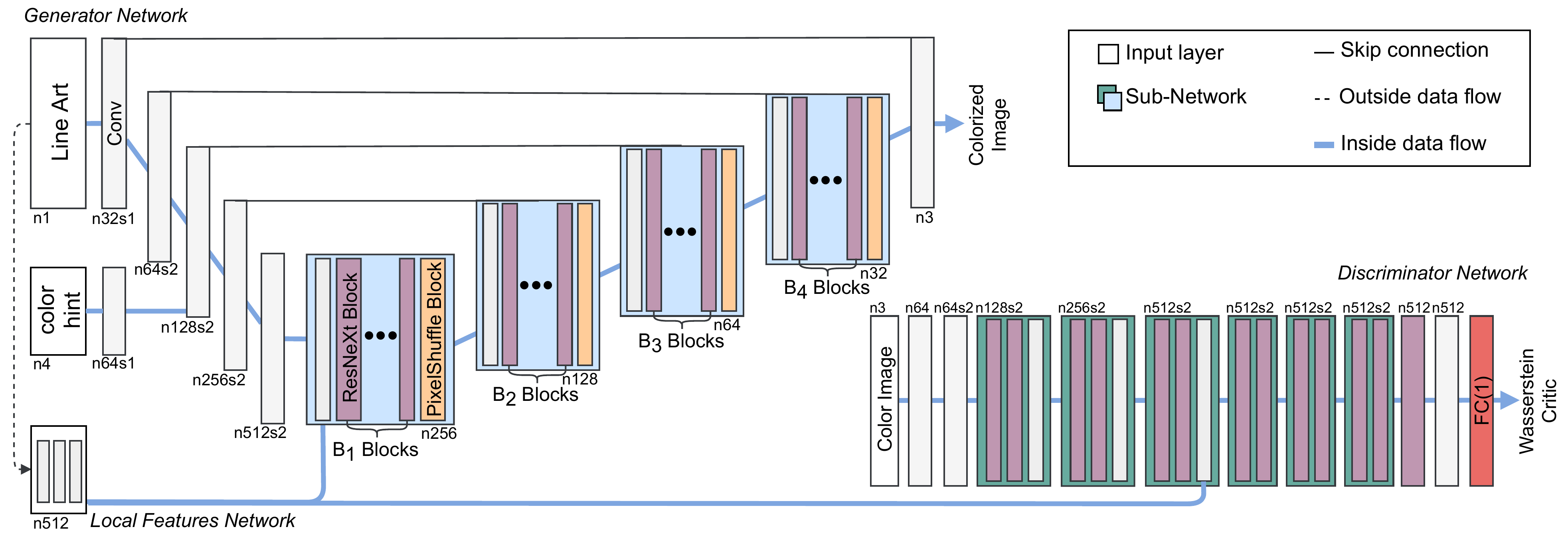}
	\caption{Architecture of Generator and Discriminator Network with corresponding number of feature maps (n) and stride (s) indicated for each convolutional block.}
	\label{pic:generator}
\end{figure*}
\subsection{Learning Framework for Colorization}
\label{sec:learn}
\begin{figure}[t]
	\centering
	\includegraphics[width=0.48\textwidth]{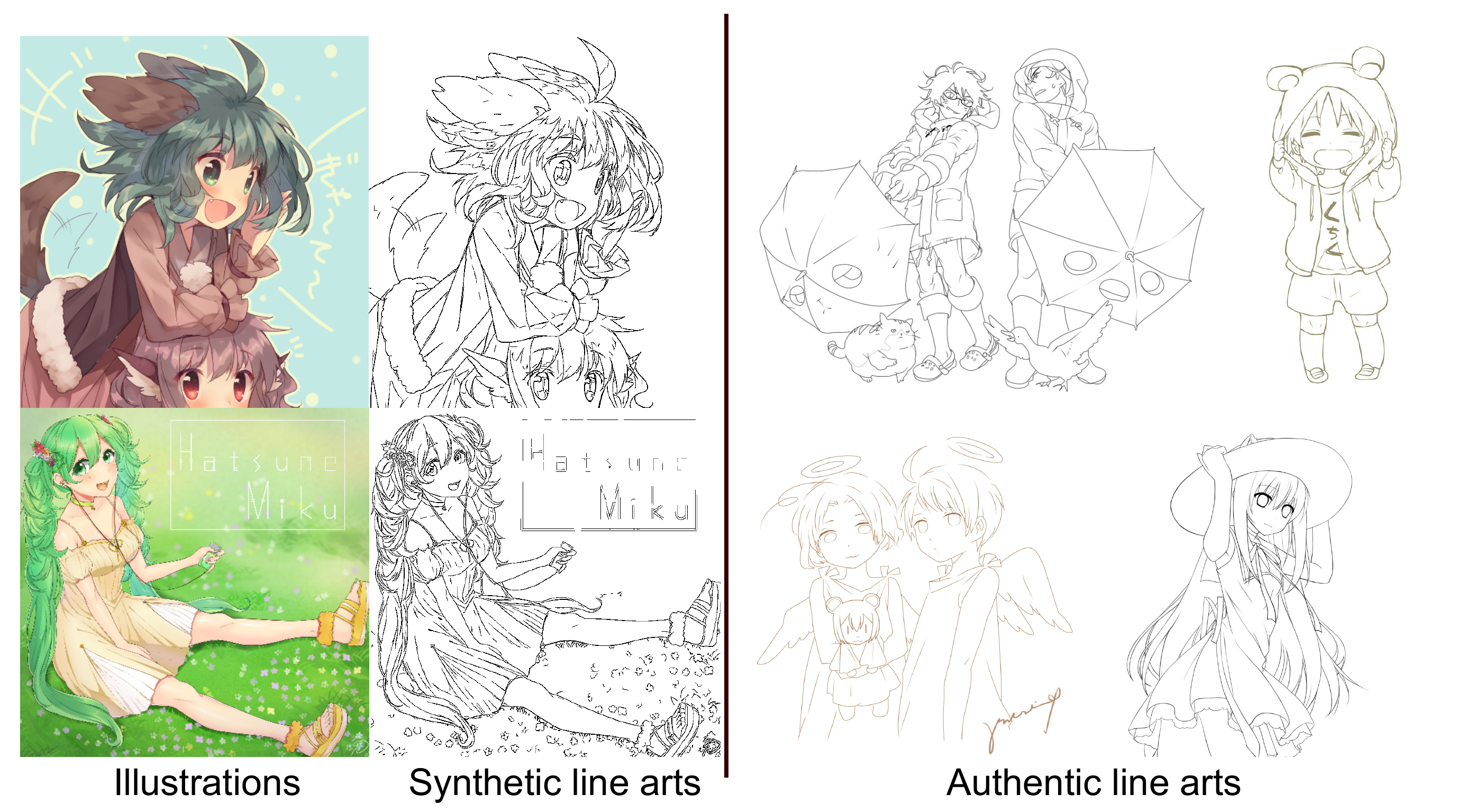}
	\caption{Sample images from our illustration dataset and authentic line art dataset, with a matching generated line art with XDoG \cite{Winnem2012XDoG} algorithm from illustrations.}
	\label{fig:data}
\end{figure}

The first input to our system is a greyscale line art image $X \in \mathbb{R}^{1 \times H \times W}$, which is a sparse,  binary image-like tensor synthesized from real illustration $Y$ with boundary detection filter XDoG \cite{Winnem2012XDoG} as shown in Figure \ref{fig:data}.

Real-world anime line arts contains a large variety of contents and are usually drawn in different styles.  It's crucial to identify the boundary of different objects and further extract semantic labels from the plain line art as the two plays an important role in generating high-quality results in image-to-image translation tasks \cite{wang2017high}.  Recent work \cite{hensman2017cgan} adopted \textit{trapped-ball segmentation} on greyscale manga images and use the segmentation to refine the cGAN colorization output, while \cite{furusawa2017comicolorization} added an extra global features network (trained to predict characters' name) to extract global feature vectors from greyscale manga images to the generator.

By extracting features from an earlier stage of a pretrained network, we introduce a local features network $\mathcal{F}_{1} $ (trained to tag illustrations) to extract semantic feature maps that contains both semantic information and spacial information directly from the line arts to the generator. We also take the local features as the conditional input of the discriminator as shown in Figure \ref{fig:overview}.  This relieves the overfitting problem since the characteristics of the man-made line arts could be very different from the synthetic line arts generated by algorithms, while the local features network is trained separately and is not affected by those synthetic line arts.  Moreover, compared with global features network, local features network preserves spatial information for the abstracted features and keeps the generator fully convolutional for arbitrary input size.

Specifically, we use the ReLU activations of the 6th convolution layer of the Illustration2Vec \cite{Saito2015Illustration2Vec} network $\mathcal{F}_{1}(X) \in \mathbb{R}^{512 \times (H/16) \times (W/16)}$ as the local feature, of which is pretrained on 1,287,596 illustrations (colored images and line arts included) predicting 1,539 labels.

The second input to the system is the simulated user hint $H$ . We sample random pixels from 4 times downsampled $Y$ as $Y_{down}$. The locations of the sampled pixels are selected by binary mask $B = R > \vert \xi \vert $ where $R\in \mathbb{R}^{1 \times (H/4) \times (W/4)}$ and $\forall r \in R, r \sim U\left( 0, 1\right)$ and we let $ \xi \sim  N\left( 1, 0.005\right)$. Together with $Y$, the tensors form color hint $H = \{Y_{down} * B, B\} \in \mathbb{R}^{4 \times (H/4) \times (W/4)}$.

The output of the system is $\mathcal{G}(X) \in \mathbb{R}^{3 \times H \times W}$, the estimate of the $RGB$ channels of the line art. The mapping is learned with a generator $\mathcal{G}$, parameterized by $\theta$, with the network architecture specified in Section \ref{sec:network} and shown in Figure \ref{pic:generator}.  We train the network to minimize the objective function in Equation \ref{equ:loss}, across $\mathcal{I}$, which represents a dataset of illustrations, line arts, color hints, and desired output colorization. Loss function $\mathcal{L}_{\mathcal{G}}$ describes how close the network output is to the ground truth.
\begin{equation}
\label{equ:loss}
\theta^{*} = \mathop{\arg\min}_{\theta} \mathbb{E}_{X,H,Y \sim \mathcal{I}} [\mathcal{L}_{\mathcal{G}} ( \mathcal{G} ( X,H,\mathcal{F}_{1}(X);\theta ), Y ) ].
\end{equation}

\subsection{Loss Function}

\label{sec:loss}

We formulate the loss function for generator $\mathcal{G}$ as a combination of content and adversarial loss:
\begin{equation}
\mathcal{L}_{\mathcal{G}} =\underbrace{\underbrace{\mathcal{L}_{cont}}_{content\ loss} + \underbrace{\lambda_{1} \cdot \mathcal{L}_{adv}}_{adv\ loss}}_{total\ loss}
\end{equation}
where the $\lambda_{1}$ equals to 1e-4 in all experiments. Similar to Isola et
al. \cite{isola2017image}, our discriminator is also conditioned, but with local features $\mathcal{F}_{1}(X)$ as conditional input and WGAN-GP \cite{gulrajani2017improved} as the critic function that distinguish between real and fake training pairs. The critic does not output a probability and the loss is calculated as the
following:
\begin{equation}
\mathcal{L}_{\mathit{adv}} =-\mathop{\mathbb{E}}_{\mathcal{G}(X,H,\mathcal{F}_{1}(X))\sim\mathbb{P}_{g}}[\mathcal{D}(\mathcal{G}(X,H,\mathcal{F}_{1}(X)), \mathcal{F}_{1}(X))].
\end{equation}
where the output of $\mathcal{F}_{1}$ denotes the feature maps obtained by a pretrained network as is described in Section \ref{sec:learn}.
To penalize color/structural mismatch between the  output of generator and ground truth,  we adopted perceptual loss \cite{johnson2016perceptual} as our content loss. Perceptual loss is a simple L2-loss based on the difference of the generated and target image CNN feature maps. It is defined as following:
\begin{equation}
\mathcal{L}_{\mathit{cont}} = \dfrac{1}{c h w}\parallel\mathcal{F}_{2}(\mathcal{G}(X,H,\mathcal{F}_{1}(X)))-\mathcal{F}_{2}(Y)\parallel_{2}^{2}.
\end{equation}
Here $c$, $h$, $w$ denotes the number of channels, height and width of the feature maps. The output of $\mathcal{F}_{2}$ denotes the feature maps obtained by the 4th convolution layer (after activation) within the VGG16 network, pretrained on ImageNet \cite{deng2009imagenet}. 

The loss definition of our discriminator $\mathcal{D}$ is formulated as a combination of wasserstein critic loss and penalty loss:
\begin{equation}
\mathcal{L}_{\mathcal{D}}=\underbrace{\underbrace{\mathcal{L}_{w}}_{critic\ loss} + \underbrace{\mathcal{L}_{p}}_{penalty\ loss}}_{total\ loss}
\end{equation}
while the critic loss is simply the WGAN \cite{arjovsky2017wasserstein} with conditional input:
\begin{equation}
\begin{aligned}
\mathcal{L}_{w}=\mathop{\mathbb{E}}_{\mathcal{G}(X,H,\mathcal{F}_{1}(X))\sim\mathbb{P}_{g}}&[\mathcal{D}(\mathcal{G}(X,H,\mathcal{F}_{1}(X)), \mathcal{F}_{1}(X))]- \\
\mathop{\mathbb{E}}_{Y\sim\mathbb{P}_{r}}&[\mathcal{D}(Y,\mathcal{F}_{1}(X))]
\end{aligned}
\end{equation}
For the penalty term, we combine the gradient penalty \cite{gulrajani2017improved} and an extra constraint term introduced by karras et al. \cite{karras2017progressive}:   
\begin{equation}
\begin{aligned}
\mathcal{L}_{p} = \ &\lambda_{2}\mathop{\mathbb{E}}_{\hat{Y}\sim\mathbb{P}_{i}}[ (\parallel\nabla_{\hat{Y}}\mathcal{D}(\hat{Y},\mathcal{F}_{1}(X))\parallel_{2}-1)^{2}] \ + \\ &\epsilon_{\mathit{drift}}\mathop{\mathbb{E}}_{Y\sim\mathbb{P}_{r}}[\mathcal{D}(Y,\mathcal{F}_{1}(X))^{2}]
\end{aligned}
\end{equation}
to keep the output value from drifting too far from zero, as well as enable us to alternate between updating the generator and discriminator on a per-minibatch basis, which reduces the training time compared to traditional setup that updates discriminator five times for every generator update. We set $\lambda_{2}=10$, $\epsilon_{\mathit{drift}}=1e-3$ in all experiments. The distribution of interpolate points $\mathbb{P}_{i}$ at which to penalize the gradient is implicitly defined as following:

\begin{equation}
\hat{Y}= \epsilon \mathcal{G}(X,H,\mathcal{F}_{1}(X)) + (1-\epsilon) Y,     \epsilon \sim U[0,1]
\end{equation}
By this we penalize the gradient over straight lines between points in the illustration distribution $\mathbb{P}_{r}$ and generator distribution $\mathbb{P}_{g}$.

\subsection{Network Architecture}
\label{sec:network}
\begin{figure*}[ht]
	\centering
	\includegraphics[width=1\textwidth]{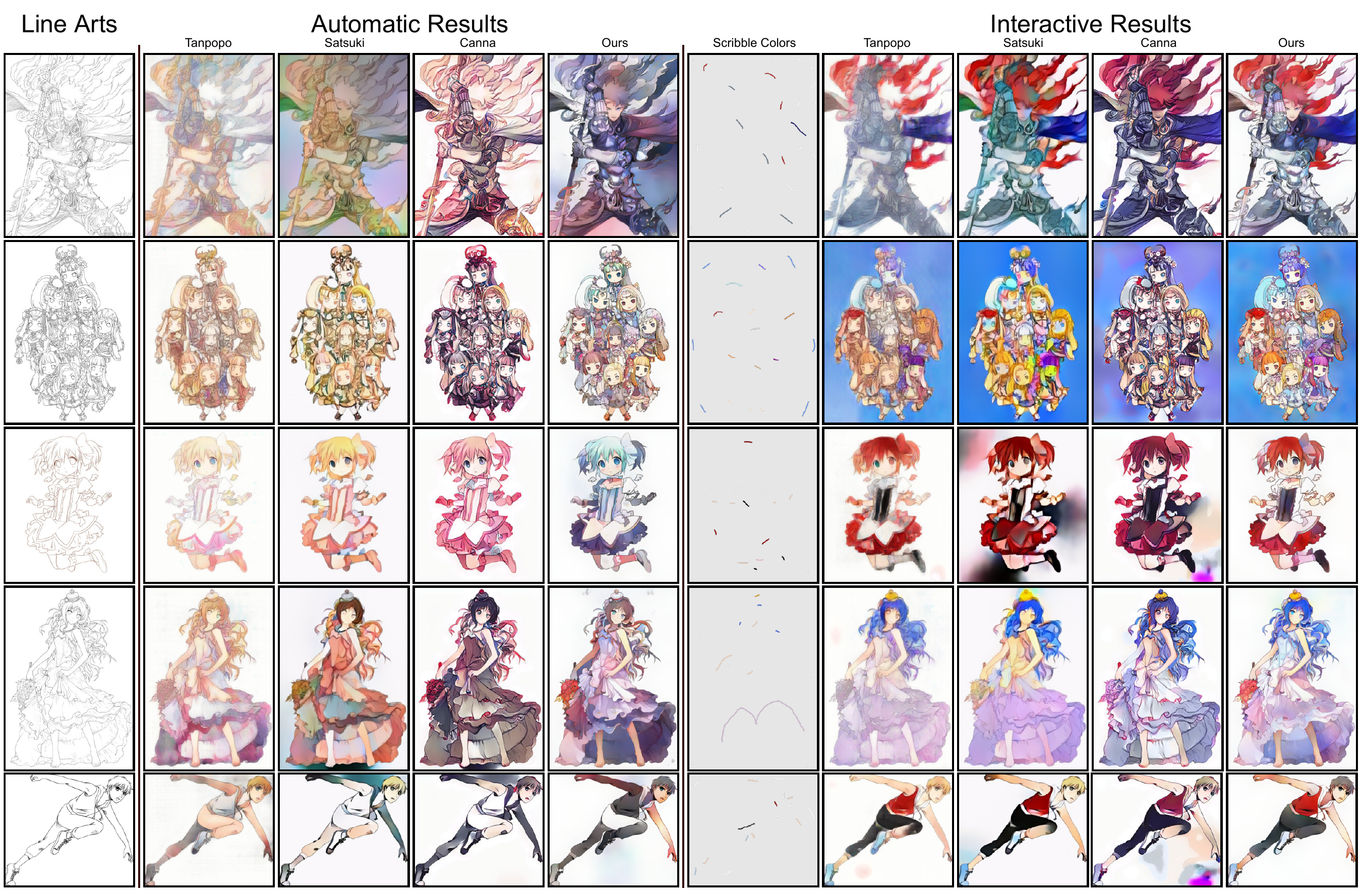}
	\caption{Scribble color-based line art colorization of authentic line arts. The first column shows the line art input image. Columns 2-5 show automatic results from three models of \cite{paintschainer} as well as ours without user input. Column 6 shows input scribble colors (generated on PaintsChainer\cite{paintschainer}, best viewed with grey background). Columns 7-10 show the results from the \cite{paintschainer} and our model, incorporating user inputs. In the selected examples in row 1, 3, 5, our system produces higher quality colorization results given variation inputs. Row 2, 4 show some failures of our model. In row 2, thew background color by the user is not successfully propagated smoothly to the image. In row4, the automatic result produce undesired gridding artifacts where the input is sparse. Images are from our line art dataset.  All the results of \cite{paintschainer} are obtained in March 2018, images are from our line art dataset. }
	\label{pic:compare}
	
\end{figure*}
For the main branch of our generator $\mathcal{G}$ which is shown in Figure \ref{pic:generator}, we employ an U-Net \cite{ronneberger2015u} architecture which has recently been used on a variety of image-to-image translation tasks\cite{isola2017image, zhu2017unpaired, wang2017high, zhang2017real}. At the front of the network locates two convolution blocks and a local features network that transform image/color hint inputs to feature maps.
%Each block contains a conv-leaky relu pair.  
Then the features are progressively halved spatially until they reach the same scale as local features. For the second half of our U-Net follows 4 sub-networks that share a similar  structure. Each sub-network contains a convolution block in the front to fuse features from skip connection(or local features for the first sub-network), then we stack $B_{n}$ ResNeXt blocks \cite{xie2017aggregated} as the core of our sub-network. Here we use ResNeXt blocks instead of Resnet blocks because ResNeXt blocks are more effective in increasing the capacity of the network. Specifically, $B_{n}$ are set to be $\{20, 10, 10, 5\}$.  We also utilize design principles from \cite{yu2017dilated} and add dilation in the ResNeXt blocks of $B_{2}, B_{3}, B_{4}$ to further increase the receptive fields without increasing the calculation cost. Finally we increase the resolution of the features with sub-pixel convolution layers as proposed by Shi et al. \cite{shi2016real}. Inspired by Nah et al.\cite{nah2017deep} and Lim et 
al. \cite{xie2017aggregated}, we did not use any normalization layer throughout our networks to keep the range flexibility for accurate colorizing. It also reduces the memory usage, computational cost and enables a deeper structure that has receptive fields large enough to "see" the whole patch with limited computational resources. We use LeakyReLU activations with slope 0.2 for every convolution layer except the last one with tanh activation. 

During the training phase, we define a discriminator network $\mathcal{D}$ as shown in Figure \ref{pic:generator}. The architecture of $\mathcal{D}$ is similar to the setup of SRGAN \cite{ledig2016photo} with some modifications. We take local features from $\mathcal{F}_{1}$ as conditional input to form a cGAN \cite{mirza2014conditional} and employ same basic block as is used in the generator $\mathcal{G}$ without dilation. We additionally stacked more layers  so that it can process $512 \times 512$  inputs.

\subsection{User Interaction}

\label{sec:interact}
One of the most intuitive ways to control the outcome of colorization is to 'scribble' some color strokes to indicate the preferred color in a region. To train a network to recognize these control signals at test time, Sangkloy et al. \cite{sangkloy2017scribbler} synthesize color strokes for the training data. Zhang et al. \cite{zhang2017real}  suggest that randomly sampled points are good enough to simulate point-based inputs. We trade off between them and use  randomly sampled points in $4\times$ downsampled scale to simulate stroke-based inputs with the intuition that each color strokes tend to have uniform color value and dense spatial information.  The way to generate training points is described in Section \ref{sec:learn}. For the user input strokes, we downsample the stroke image to quarter resolution with max-pooling and remove half of the input pixels by setting stroke image $Y_{down}$ and binary mask $B$ to 0 with an interval of 1. This removes the redundancy of strokes as well as preserve spatial information and simulate the sparse training input. In this way, we are able to cover the input space adequately and train an effective model.

\section{Experiments}

\subsection{Dataset}
\label{sec:data}

Nico-opendata \cite{ikuta2016} and Danbooru2017 \cite{danbooru2017} provide large illustration datasets that contain illustrations and their associated metadata. But they are not adoptable for our task as they contain messy scribbles as well as sketches/line arts mixed in the dataset. These noises are hard to clean and could be harmful to the training process.

Due to the lack of publicly available quality illustration/line art dataset. We propose two quality datasets\footnote{Both datasets are available  at  https://pan.baidu.com/s/1oHBqdo2cdM8jOmAaix9xpw}: illustration dataset and line art dataset. We collected 21930 colored anime illustrations and 2779 authentic line arts from the internet for training and benchmarking.  We further apply the boundary detection filter XDoG \cite{Winnem2012XDoG} to generate synthetic sketches-drawing pairs. 

To simulate the  line drawings sketched by artists, we set the parameters of XDoG algorithm with $\varphi=1\times10^{9}$  to keep a step transition at the border of sketch lines. We randomly set $\sigma$ to be 0.3/0.4/0.5 to get different levels of line thickness thus generalize the network on various line width. Additionally, we set $\tau=0.95, \kappa=4.5$ as default. 

\subsection{Experimental Settings}	
\label{sec:Settings}

The PyTorch framework \cite{paszke2017automatic} is used to implement our model. All training was performed on a single NVIDIA GTX 1080ti GPU.
We use ADAM \cite{Kingma2014Adam} optimizer with hyperparameters $\beta_{1}=0.5, \beta_{2}=0.9$ and batch size 4 due to limited GPU memory. All networks were trained from scratch, with an initial learning rate of 1e-4 for both generator and discriminator. After 125k iterations, the learning rate is decreased to 1e-5. Total training takes 250k iterations to converge. As is described in Section \ref{sec:loss}, we perform one gradient descent step on $\mathcal{D}$ and simultaneously one step on $\mathcal{G}$.

To take non-black sketches into account, every sketch image $x$ is randomly scaled to $\hat{x}=1-\lambda \left( 1-x\right) $ , where $\lambda$ is sampled from an uniform distribution $U\left( 0.7, 1\right) $. We resize the image pairs with shortest sides to be 512 and then randomly corp to 512x512 before random horizontal flipping.

\subsection{Quantitative  Comparisons}
\label{sec:Quantitative}
Evaluating the quality of synthesized images is an open and difficult problem \cite{salimans2016improved}. Traditional metrics such as PSNR  do not assess joint statistics between targets and results. Unlike greyscale image colorization,  only few authentic line arts have corresponding  ground truths available to perform PSNR evaluation. In order to evaluate the visual quality of our results, we employ two metrics. First, we adopted Fr\'echet Inception Distance \cite{heusel2017gans}  to measure the similarity between  colorized line arts and  authentic illustrations. It is calculated by computing the Fr\'echet distance between two Gaussians fitted to feature representations of the pretrained Inception network \cite{szegedy2016rethinking}. Second, we perform a mean opinion score test to quantify the   ability of different approaches in reconstructing perceptually convincing colorization results, i.e., whether the results are plausible to a human observer.

\begin{figure*}[t]
	\centering
	\includegraphics[width=1\textwidth]{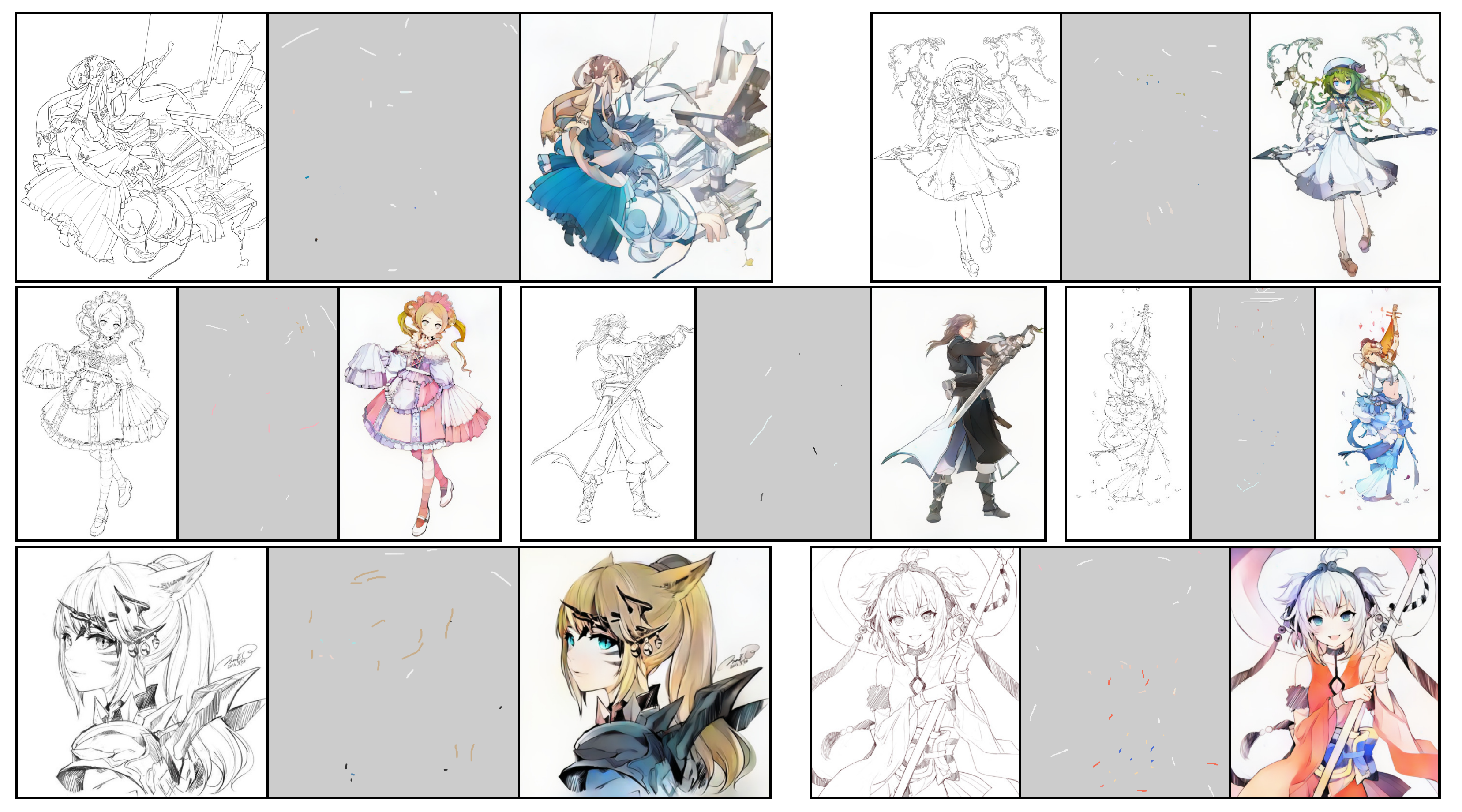}
	\caption{Selected User Study Results. We show line art images with user inputs (best viewed with grey background), along side the outputs from our method. All images was colorized by a novice user and from our line art dataset.}
	
	\label{pic:user}
\end{figure*}
\subsubsection{Fr\'echet Inception Distance (FID)}
\begin{table}[t]
	\begin{center}
		\begin{tabular}{|l|c|c|}
			\hline
			Training Configuration				& FID	\\
			\hline
			PaintsChainer(canna) 	&103.24 $\pm$ 0.18	\\
			PaintsChainer(tanpopo)		&85.38 $\pm$ 0.21		\\
			PaintsChainer(satsuki)	&81.91 $\pm$ 0.26		\\
			Ours (w/o Adversarial Loss)		&70.90 $\pm$ 0.13	\\
			Ours (w/o Local Features Network)	 		 	&60.73 $\pm$ 0.22	\\
			Ours  										&\textbf{57.06 $\pm$ 0.16}\\
			\hline
		\end{tabular}
	\end{center}
	\caption{Quantitative comparison of Fr\'echet Inception Distance without added color hint. The results are calculated between automatic colorization results of 2779 authentic line arts and 21930 illustrations from our proposed datasets.}
	\label{wow}
\end{table}
Fr\'echet Inception Distance (FID)  is adopted as our metric because it can detect intra-class mode dropping (i.e., a model generates only one image per class will have a bad FID), and can measure diversity, quality of generated samples \cite{lucic2017gans, heusel2017gans}. Intuitively a small FID indicates that the distribution of two set of images is similar. Since two of paintschainer's models \cite{paintschainer} are not open-sourced (\textit{canna}, \textit{satsuki}), it's hard to keep identical user input during testing. Therefore, to quantify the quality of the results under the same condition, we only synthesize colorized line arts automatically (i.e., without color hints)  on our authentic line art dataset for all methods and report their FID scores on our proposed illustration dataset. The obtained FID scores are reported in Table \ref{wow}. As can be seen, our final method achieves a smaller FID than other methods. Removing adversarial loss or local features network during training both leads to a higher FID. 

\subsubsection{Mean opinion score (MOS) testing}
\begin{figure}[t]
	\centering
	\includegraphics[width=0.48\textwidth]{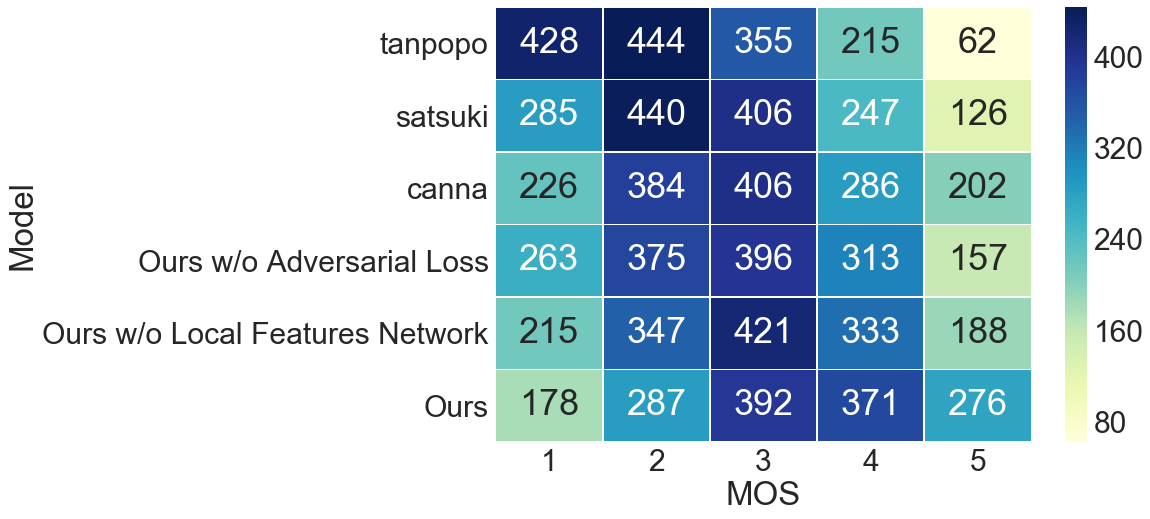}
	\caption{Color-coded distribution of MOS scores on our line art dataset. For each method 1504 samples (94 images $\times$ 16 raters) were assessed. Number of ratings annotated in the corresponding blocks.}
	\label{fig:mos}

\end{figure}

\begin{table}[t]
	\begin{center}
		\begin{tabular}{|l|c|c|}
			\hline
			Training Configuration				 & MOS	\\
			\hline
			
			PaintsChainer(canna) 		&2.903	\\
			PaintsChainer(tanpopo)			&2.361	\\
			PaintsChainer(satsuki)		&2.660	\\
			Ours (w/o Adversarial Loss)		&	2.818	\\
			Ours (w/o Local Features Network)	 		 	 &	2.955	\\
			Ours  										&\textbf{3.186}	\\					
			\hline
		\end{tabular}
	\end{center}
	\caption{Performance of different methods for automatic colorization on our line art dataset. Our method achieves significantly higher  ($p < 0.01$) MOS score than other methods.	}
	\label{woow}

\end{table}

We have performed a MOS test to quantify the ability of different approaches to reconstruct perceptually convincing colorization result. Specifically, we asked 16 raters to assign an integral score from 1 (bad quality) to 5 (excellent quality) to the automatically colorized images. The raters rated 6 versions of 1504 randomly sampled results on our line art dataset: ours, ours w/o adversarial loss, ours w/o local features network, PaintsChainer's \cite{paintschainer} model \textit{canna}, \textit{satsuki}, \textit{tanpopo}. Each rater thus rated
564 instances (6 versions of 94 line arts) that were presented in a randomized fashion.
The experimental results of the conducted MOS tests are summarized in Table \ref{woow} and Figure  \ref{fig:mos}. All raters were not calibrated and statistical tests were performed as one-tailed hypothesis test of difference in means, significance determined at $p < 0.01$ for our final method, confirming that our method outperforms all reference methods.  We noticed that method \textit{canna} \cite{paintschainer} has a better performance in terms of MOS testing than FID, we conclude that it  focuses more on generating eye pleasing results rather than results matching the authentic illustration distribution.

\subsection{Analysis of the Architectures}
\label{sec:Analysis}
\begin{figure}[t]
	\centering
	\includegraphics[width=0.48\textwidth]{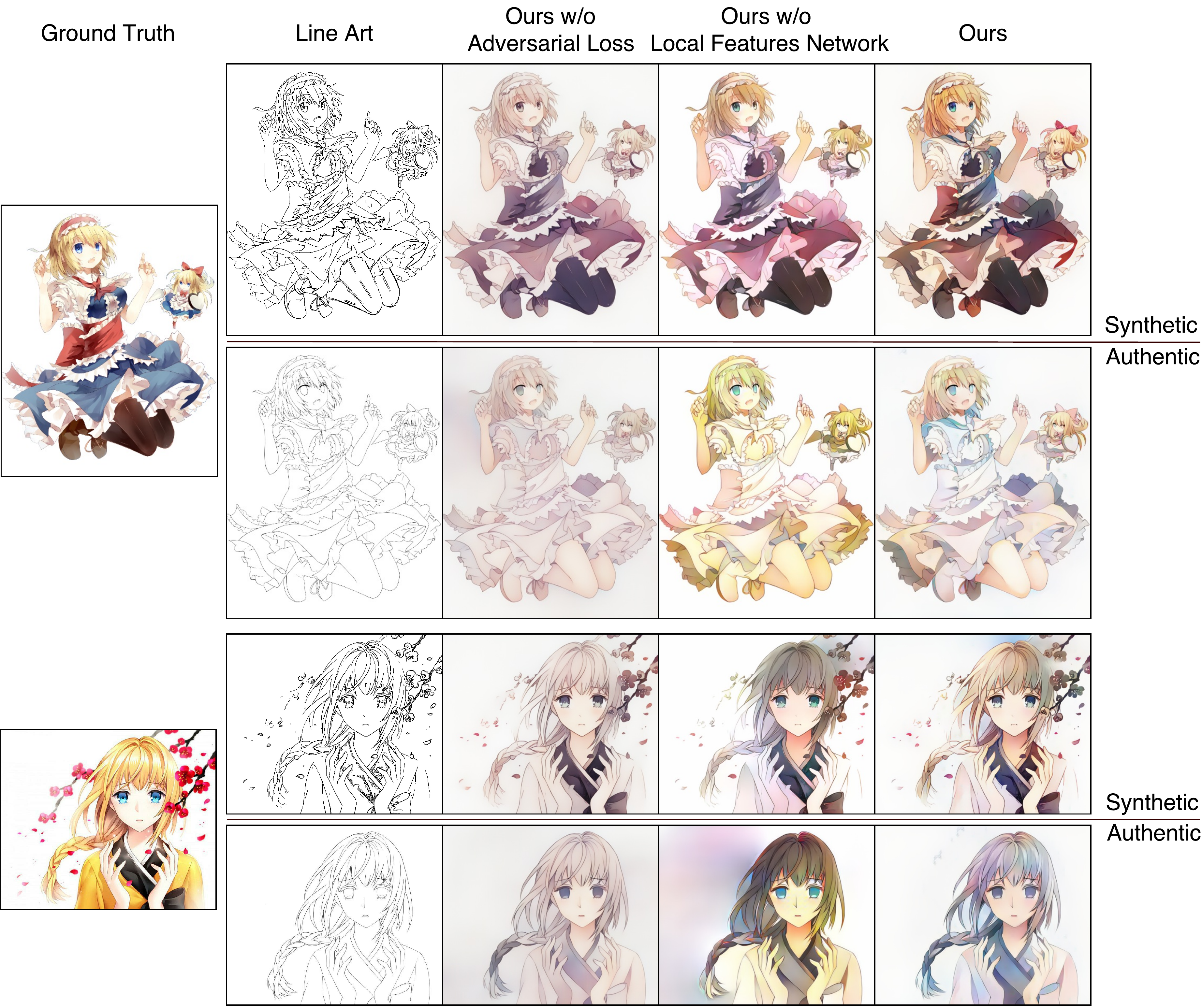}
	\caption{Comparison on the automatic colorization results from synthetic line arts and authentic line arts. 
		Results generated from synthetic line art are perceptually satisfying but overfitted over XDoG artifacts for all methods. Nevertheless, our method with local features network and adversarial loss can still generate illustration-realistic result from man-made line arts.}
	\label{fig:overfit}
\end{figure}
We investigate the value of our proposed cGAN training methodology and local features network for colorization task, and the results are shown in Figure \ref{fig:overfit}. It can be seen that method with our adversarial loss and local features network can achieve better performance than those without on "in the wild" authentic line arts. All methods can generate results with greyscale values matching the ground truth on synthetic line arts, which shows an overfitting on synthetic line art artifacts. In the case of authentic line arts, the method without adversarial loss generates results that lack saturation, which indicates the underfitting of the model. While method without local features network generates unusually colorized results, showing that the lack of generalization over authentic line arts. It is apparent that adversarial loss leads to a more sharp, illustration-realistic results, while using the local features network, on the other hand, helps to generalize the network on authentic line arts which are unseen during training.

\subsection{Color Strokes Guided Colorization}
\label{sec:exmaples}
A benefit of our system is that the network predicts user-intended actions based on learned priors. Figure \ref{pic:compare} shows side-by-side comparisons of the results against the state of the art of \cite{paintschainer} and Figure \ref{pic:user} shows example results generated by a novice user. Our method performs better at hallucinating missing details (such as shadings and color of the eyes) as well as generating diverse results given color strokes and simple line arts drawn with various  styles. It can also be seen that our network is able to propagate the input color to the relevant regions respecting object boundaries.

\section{Conclusion}

In this paper, we propose a conditional adversarial illustration synthesis architecture and a loss for stroke-based user-guided anime line art colorization.  We explicitly proposed local features network to ease the gap between synthetic data and authentic data. A novel deep cGAN network is described with specially designed sub-networks to increase the network capacity as well as receptive fields. Furthermore, we collected two datasets with quality anime illustrations and line arts, enabling efficient training and rigorous evaluation. Extensive  experiments show that our approach outperforms the state-of-the-art methods in both qualitative and quantitative ways.

%\end{document}  % This is where a 'short' article might terminate

\begin{acks}
	This work was supported in part by the National Natural Science Foundation of China (NSFC)  under Grants  No.~\grantnum{}{61472059}  and  No.~\grantnum{}{61772108}.
	
\end{acks}

\bibliographystyle{ACM-Reference-Format}
\balance 
\bibliography{sample-bibliography}

\end{document}